%% file: main.tex
\documentclass{article}
\usepackage[preprint]{corl_2026}         
\usepackage{times}
\usepackage{amsmath,amssymb,amsfonts,bm}
\usepackage{graphicx}
\usepackage{xcolor}
\usepackage{booktabs}
\usepackage{multirow}
\usepackage{array}
\usepackage{makecell}
\usepackage{enumitem}
\usepackage{csquotes}
\usepackage{tikz}
\usetikzlibrary{arrows.meta, calc, positioning, shapes.geometric,
                shapes.symbols, decorations.pathmorphing, backgrounds, fit, shadows}
\usepackage{pgfplots}
\pgfplotsset{compat=1.18}
\usepackage{xspace}
\newif\ifappendixonly
\appendixonlyfalse    
\newcommand{\bodyref}[2]{\ifappendixonly #2~of the main paper\else\ref{#1}\fi}
\newcommand{\blfootnote}[1]{\begingroup\renewcommand\thefootnote{}\footnotetext{#1}\endgroup}
\usepackage{silence}
\makeatletter
\renewcommand{\@maketitle}{%
  \vbox{%
    \hsize\textwidth
    \linewidth\hsize
    \vskip 0.1in
    \centering
    {\Large\bf \@title\par}%
    \if@conferencefinal
      \def\And{\hskip 1em\relax}%
      \def\AND{\hfil\linebreak[4]\hfil}%
      \begin{tabular}[t]{c}\bf\rule{\z@}{18\p@}\@author\end{tabular}%
    \else
      \if@preprinttype
        \def\And{\hskip 1em\relax}%
        \def\AND{\hfil\linebreak[4]\hfil}%
        \begin{tabular}[t]{c}\bf\rule{\z@}{18\p@}\@author\end{tabular}%
      \else
        \begin{tabular}[t]{c}\bf\rule{\z@}{24\p@}
          Anonymous Author(s) \\ Affiliation \\ Address \\ \texttt{email} \\
        \end{tabular}%
      \fi
    \fi
    \vskip 0.3in \@minus 0.1in
  }
}
\makeatother
\definecolor{accent}{RGB}{40,40,45}      
\definecolor{accent3}{RGB}{178,34,34}    
\definecolor{accent4}{RGB}{110,110,110}  
\definecolor{codepink}{RGB}{233,30,99}   
\newcommand{\method}{\textsc{Semantic Flip}\xspace}
\newcommand{\qflip}{\textsc{Q-Flip}\xspace}
\newcommand{\vflip}{\textsc{V-Flip}\xspace}
\newcommand{\spacereject}{\textsc{SpaceReject}\xspace}
\newcommand{\abstaineqa}{\textsc{AbstainEQA}\xspace}
\newcommand{\metamemory}{\textsc{Meta-Memory}\xspace}
\newcommand{\spacelocqa}{\textsc{SpaceLocQA}\xspace}
\title{\textsc{Semantic Flip}: Synthetic OOD Generation for Robust Refusal\\
       in Embodied Question Answering and Spatial Localization}
\author{
  Dongbin Na\textsuperscript{*,\dag} \quad Chanwoo Kim\textsuperscript{*} \quad Giyun Choi \quad Dooyoung Hong\textsuperscript{\dag} \\
  RGA Inc. \\
  \texttt{\{dongbinna, cwkim, cky, dooyoung\}@rgarobot.com}
}
\begin{document}
\ifappendixonly
  \thispagestyle{plain}%
  \begin{center}{\large\bfseries Supplementary Material}\end{center}
  \vspace{1.0em}
\else
  \maketitle  
  \blfootnote{\textsuperscript{*}Equal contribution.}%
  \blfootnote{\textsuperscript{\dag}Correspondence to \texttt{dongbinna@postech.ac.kr} and \texttt{dooyoung@rgarobot.com}.}%
  \blfootnote{The project page: \textcolor{codepink}{\texttt{https://ndb796.github.io/SemanticFlip}}}%
\fi
\ifappendixonly\else  
\begin{abstract}
Detecting unanswerable user queries remains essential for the reliable deployment of real-world embodied agents. 
However, modern vision-language models (VLMs) often generate overly confident answers even when the available visual memory cannot support the query.
Such overconfidence poses various task-dependent risks. 
The agent may provide misleading information to the user in Embodied Question Answering and select an arbitrary coordinate and physically guide the user there in spatial reasoning for navigation.
Despite these high stakes, only a few prior studies directly address when and how an embodied VLM should respond with \enquote{I do not know.}
This work proposes \method{}, a simple yet effective framework that synthesizes auxiliary out-of-distribution (OOD) samples for embodied refusal without requiring external OOD annotations.
The key idea is to independently transform the query and video memory to construct auxiliary OOD pairs that lack sufficient visual grounding.
These synthesized pairs enable training a lightweight rejection module on top of a frozen pretrained VLM.
The module attaches to any existing VLM-based pipeline without retraining the underlying model.
Across two complementary benchmarks, \method{} consistently outperforms strong prompting baselines. 
This work also introduces \spacereject{}, a new refusal benchmark for spatial localization with deliberately unanswerable queries over long video memory, where \method{} achieves an $F_1$ score of $0.9559$.
The source codes and datasets are publicly available at \textcolor{codepink}{\texttt{https://github.com/ndb796/SemanticFlip}}.
\end{abstract}
\section{Introduction}
\label{sec:intro}
Mobile delivery robots receive natural-language queries at deployment time, yet many real user queries are intrinsically unanswerable.
Users may ask about unseen objects, ambiguous referents, false premises, or subjective judgments~\citep{wu2025abstaineqa}.
An embodied agent should avoid hallucinated responses by honestly saying \enquote{I do not know} or initiating a multi-turn clarification dialogue~\citep{majumdar2024openeqa}.
Existing embodied-agent pipelines still do not treat refusal as a first-class behavior.
Instead, the model typically commits to the most plausible answer and presents it confidently.
This behavior carries different costs across tasks.
Users can often correct misleading information in Embodied Question Answering (EQA) through follow-up dialogue, whereas failures in spatial reasoning for physical robot control carry much higher costs.
State-of-the-art memory-augmented agents such as ReMEmbR and \metamemory{} directly emit an $(x,y)$ navigation target and produce \emph{some} coordinate even for unanswerable queries~\cite{anwar2024remembr, mao2025metamemory}.
Their large mean Euclidean errors on NaVQA, $28.5$~m for ReMEmbR and $21.7$~m for \metamemory{}, indicate that a robot may physically navigate to an essentially arbitrary point.
A robust abstention module, therefore, prevents the most harmful failures in both EQA and spatial navigation.
Existing approaches do not fully solve this problem.
Prompt-based methods can encourage \enquote{I do not know}, but they rely on surface-level instructions and remain sensitive to prompt wording~\cite{wu2025abstaineqa}.
Supervised fine-tuning trains abstention more directly, but it assumes known OOD categories and curated representative examples in advance, which conflicts with the open-ended nature of deployment-time unanswerable queries.
Fine-tuned models may therefore learn dataset-specific shortcuts rather than grounded unanswerability, consistent with prior findings that a simple TF-IDF classifier can match a fine-tuned model.
CoT prompting also fails to provide a reliable path to abstention because it tends to construct plausible answer paths even when visual evidence supports no valid answer~\cite{wei2022cot}.
Recent work further shows that reasoning-oriented fine-tuning can degrade abstention rather than improve it~\cite{kirichenko2025abstentionbench}.
These limitations motivate a training signal that directly captures ungroundability without manually curated OOD annotations.
This work proposes \method{}, a simple and effective framework that synthesizes auxiliary OOD samples for embodied refusal without external OOD annotation.
The key idea corrupts one modality, either the video memory or the text query, while keeping the other intact, producing OOD pairs that lack valid cross-modal grounding.
Built on the same frozen VLM as the prompting baselines, \method{} adds only a lightweight rejection module.
On \abstaineqa{}, \method{} reaches an $F_1$ score of $0.7110$ with a frozen $7$B VLM, outperforming the strongest $32$B prompting baseline at $0.6746$.
On \spacereject{}, \method{} further achieves an $F_1$ score of $0.9559$, showing strong refusal performance in spatial localization over long video memory.
These results show that synthetic OOD supervision can make a practical open-source $7$B backbone outperform larger prompt-only alternatives.
\paragraph{Technical contributions.}
\begin{enumerate}[itemsep=0pt,topsep=2pt]
    \item This work introduces \method{}, a synthetic OOD generation framework for embodied refusal that creates auxiliary unanswerable pairs by independently corrupting either the query (\qflip{}) or the video memory (\vflip{}) without external OOD annotation, and attaches to frozen VLM-based embodied pipelines as a lightweight rejection module.
    \item \method{} outperforms strong prompting baselines across two complementary benchmarks. On \abstaineqa{}, it reaches an $F_1$ score of $0.7110$ with a frozen $7$B VLM, surpassing the strongest $32$B prompting baseline at $0.6746$. On \spacereject{}, it achieves an $F_1$ score of $0.9559$.
    \item This work introduces \spacereject{}, a spatial-localization refusal benchmark that extends \spacelocqa{} with $135$ unanswerable queries over long video memory, and additionally releases \textsc{SpaceRejectExtra}, a larger extension with $2{,}520$ newly curated OOD queries across six sequences.
\end{enumerate}
\section{Related Work}
\label{sec:related}
\paragraph{Embodied question answering and refusal.}
\citet{das2018embodied} introduced EQA, which OpenEQA later extended to foundation-model settings~\cite{majumdar2024openeqa}.
Most EQA benchmarks assume answerable questions grounded in the observed environment.
\abstaineqa{} defines a new setting that includes unanswerable questions across five categories, namely \textit{Actionability Limitation}, \textit{Referential Underspecification}, \textit{Preference Dependence}, \textit{Information Unavailability}, and \textit{False Presupposition}~\cite{wu2025abstaineqa}.
Interestingly, its best fine-tuned model performs similarly to a text-only classifier, which suggests that the previous work could not fully leverage vision-side supervision for reliable embodied refusal.
\paragraph{Memory-augmented robot navigation and spatial QA.}
ReMEmbR introduces retrieval-augmented memory for long-horizon navigation and presents the NaVQA benchmark~\cite{anwar2024remembr}.
\metamemory{} has formalized Spatial Localization QA (SLQA) and has proposed a three-tool LLM agent that combines semantic-similarity retrieval, spatial-range retrieval, and memory integration~\cite{mao2025metamemory}.
Both benchmarks assume that each query has a valid spatial answer.
\paragraph{Out-of-distribution detection and outlier exposure.}
Prior studies have examined OOD detection using post hoc scoring and ensemble-based uncertainty estimates~\cite{hendrycks2017msp,liang2018odin,lee2018mahalanobis,lakshminarayanan2017deepensembles}.
These methods do not directly fit embodied EQA because the task takes a multimodal video-query pair as input and produces a free-form textual answer rather than a label over a fixed class set.
Classical class-conditional scores such as MSP and ODIN, therefore, do not transfer cleanly to this setting~\cite{hendrycks2017msp,liang2018odin}.
Outlier exposure (OE) provides a closer analogue because it trains a classifier with auxiliary OOD data\citep{hendrycks2019oe}.
However, OE requires large-scale auxiliary OOD data, which deployment settings often lack.
Recent studies address this limitation by synthesizing surrogate OOD samples from ID data through token-level masking, input mixup, and marginal-feature disentangling~\cite{kim2023poe,choi2024mim,kim2023keyfeature}.
\paragraph{Selective prediction and calibration.}
The selective-prediction literature formalizes abstention under input uncertainty~\cite{elyaniv2010foundations,geifman2017selectivenet}.
Prior studies show that modern neural networks often suffer from miscalibration, and recent studies further report that LLM self-reports of uncertainty remain unreliable~\cite{guo2017calibration,kadavath2022know, yin2023didask}.
\paragraph{Open-source vision-language components.}
Open-source VLMs expose the multimodal representations on which downstream rejection modules can operate, which makes embodied refusal more reproducible and inspectable than reliance on opaque confidence from commercial APIs~\cite{bai2025qwen25vl,liu2023llava,radford2021clip}.
Open-source tools for open-vocabulary detection, inpainting, and parsing further support controlled perturbations of visual memory and language queries~\cite{liu2024groundingdino,suvorov2022lama,spacy}.
\paragraph{Chain-of-thought and refusal.}
CoT prompting has recently been adopted across many downstream tasks because it often improves reasoning by encouraging intermediate steps~\cite{wei2022cot,kojima2022zerocot}. 
However, refusal requires a model to reject unsupported answer paths rather than elaborate a plausible one, and prior calibration studies report that CoT can increase overconfidence in incorrect answers.
\section{\method{}}
\label{sec:method}
\begin{figure}[ht]
    \centering
    \includegraphics[width=\textwidth]{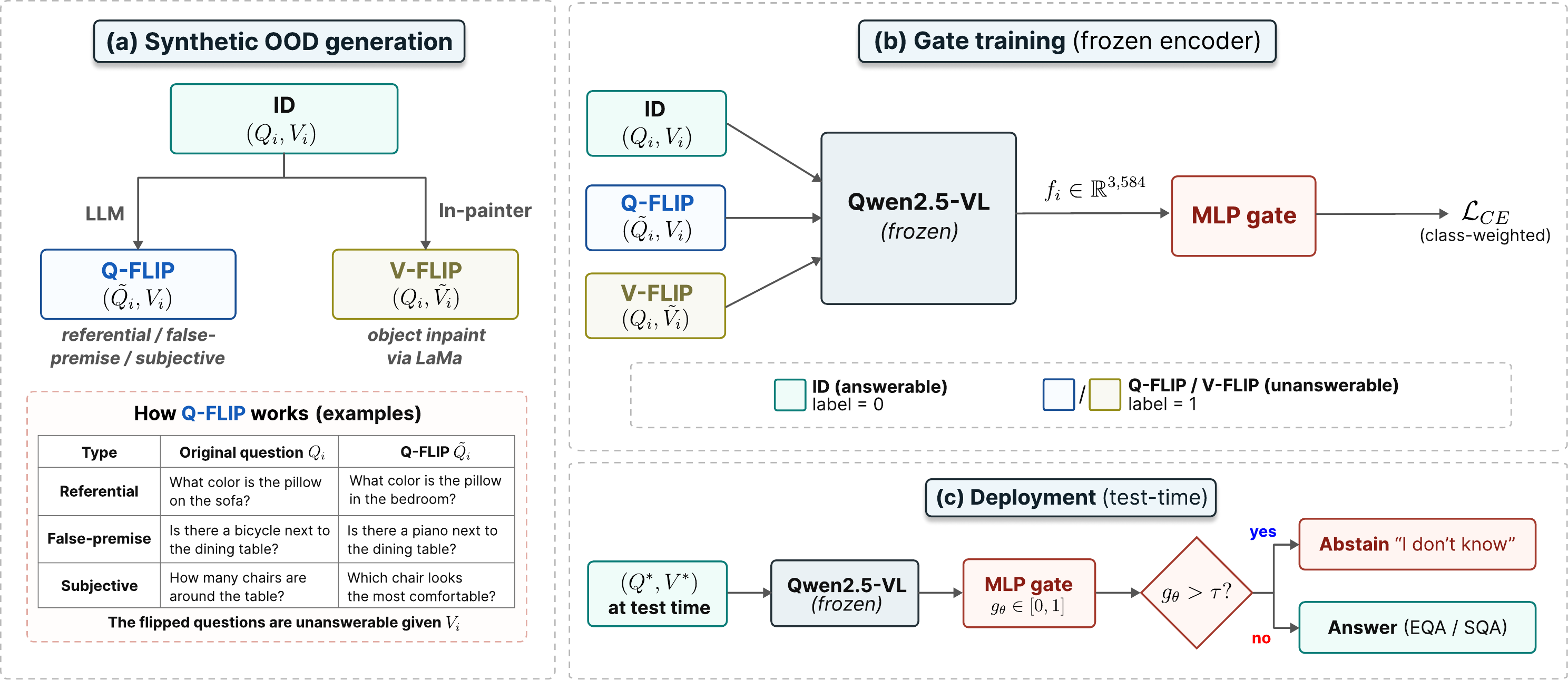}
    \caption{\textbf{\method{} overview.} (a) From in-distribution training pairs,
\qflip{} corrupts the query while keeping the video, and \vflip{} corrupts the
video while keeping the query. (b) A frozen VLM encoder produces joint embeddings
for all three distributions, and only a lightweight rejection module is trained.
(c) At test time, the module plugs into both an EQA decoder and a memory-augmented
navigation agent.}
    \label{fig:overview}
\end{figure}
This work presents \method{}, a framework for embodied refusal that synthesizes two complementary OOD distributions from in-distribution training pairs. 
\qflip{} corrupts the query while keeping the video memory, whereas \vflip{} corrupts the video memory while keeping the query.
A frozen VLM encoder produces joint embeddings for ID, \qflip{}, and \vflip{} samples, on top of which a lightweight MLP rejection gate is trained.
At test time, the gate plugs into both an EQA decoder (rejecting unanswerable questions) and a memory-augmented navigation agent (vetoing unsupported coordinate proposals before the robot moves).
\subsection{Problem formulation}
The proposed method considers an embodied agent that answers a natural-language query $Q$ using a visual memory $V$.
The training set contains only answerable examples $\mathcal{D}_{\text{ID}} = \{(Q_i, V_i, A_i)\}_{i=1}^{N}$, where $A_i$ denotes an answer grounded in $V_i$.
At deployment time, however, the agent may receive queries for which the available visual memory does not provide sufficient evidence to determine a valid answer.
The goal is to learn a rejection module $g_\theta : (Q,V) \mapsto [0,1]$, where low scores indicate answerable pairs and high scores indicate unanswerable pairs.
The desired behavior is $g_\theta(Q_i,V_i) \approx 0$ for answerable training pairs $(Q_i,V_i) \in \mathcal{D}_{\text{ID}}$, and $g_\theta(Q,V) \approx 1$ for unanswerable deployment inputs.
The main challenge is that the training data contains no manually labelled unanswerable pairs.
\method{} addresses this challenge by synthesizing auxiliary unanswerable pairs from answerable ones.
The agent should answer only when the available visual memory provides sufficient evidence to ground the language query.
If the query asks for unsupported information, or if the memory no longer contains the required referent, the pair should trigger abstention.
Let $\mathcal{T}_Q$ and $\mathcal{T}_V$ denote query-side and video-side corruption operators.
Given an answerable pair $(Q_i,V_i)$, \method{} constructs
\begin{equation}
\mathcal{D}_{\qflip{}} =
\bigl\{(\mathcal{T}_Q(Q_i), V_i)\bigr\}_{i},
\qquad
\mathcal{D}_{\vflip{}} =
\bigl\{(Q_i, \mathcal{T}_V(V_i))\bigr\}_{i}.
\end{equation}
Q-Flip changes the query while preserving the visual memory, creating queries with insufficient visual grounding, such as ambiguous referents, false premises, or subjective requests.
V-Flip preserves the query while removing the target referent from the visual memory through a \texttt{parse} $\rightarrow$ \texttt{detect} $\rightarrow$ \texttt{inpaint} pipeline.
\begin{figure}[ht]
\centering
\includegraphics[width=1.0\textwidth]{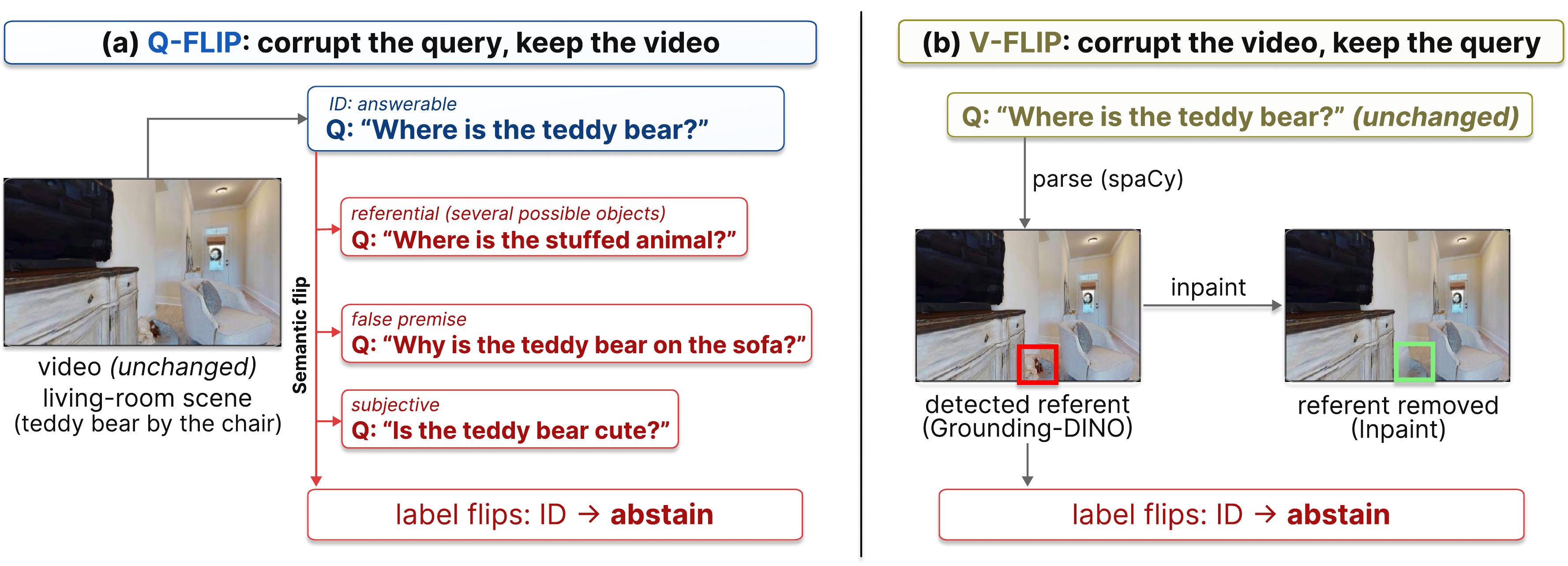}
\caption{\textbf{Concrete examples of \qflip{} and \vflip{}.} \qflip{} keeps the
video and rewrites the query into an ungroundable variant, such as a referential,
false premise, or subjective question, while \vflip{} keeps the query and erases its referent
from every frame so that the label flips to \emph{abstain}.}
\label{fig:flip-examples}
\end{figure}
Both transformations change only one axis of an originally answerable pair, as shown in Figure~\ref{fig:flip-examples}.
This minimal corruption keeps the other axis in-distribution and discourages the rejection module from relying on single-modality artifacts.
Both \qflip{} and \vflip{} samples receive the label \emph{abstain}.
\subsection{\qflip{} for linguistic corruption}
\label{sec:qflip}
\qflip{} keeps the visual memory $V_i$ unchanged and corrupts only the query for each ID example $(Q_i, V_i)$.
It generates a new query $\tilde{Q}_i$ such that $(\tilde{Q}_i, V_i)$ no longer has sufficient visual grounding.
The coarse variant asks an LLM to write a question that the given video cannot answer without using any abstention taxonomy.
The fine variant uses three \abstaineqa{} categories, namely referential underspecification, false premise, and subjective judgment, and thus assumes partial prior knowledge of abstention types.
It excludes \textit{Actionability Limitation} and \textit{Information Unavailability} during synthesis to evaluate generalization to unseen abstention categories.
Appendix~A provides the full prompts, quality filters, and coarse-versus-fine \qflip{} comparison.
\subsection{\vflip{} for perceptual corruption}
\label{sec:vflip}
\vflip{} keeps the query $Q_i$ unchanged and corrupts only the visual memory for each ID example $(Q_i, V_i)$, creating $\tilde{V}_i$ in which the original query can no longer be grounded.
Figure~\ref{fig:flip-examples}~(b) illustrates the pipeline.
spaCy extracts the target noun phrase~\cite{spacy}, Grounding-DINO localizes it across video frames~\cite{liu2024groundingdino}, and LaMa removes the localized regions through inpainting~\cite{suvorov2022lama}.
The resulting pair $(Q_i, \tilde{V}_i)$ receives the label \emph{abstain} because the visual referent required by the query no longer appears in the memory.
Although inpainting may leave residual artifacts, \vflip{} still removes the direct visual evidence required by the original query and provides auxiliary OOD supervision.
Appendix~C compares inpainting with simpler perceptual corruptions, including \textit{dilation}, \textit{graying}, \textit{noise}, and \textit{mean-color fill}.
\subsection{Frozen-encoder rejection module}
\label{sec:gate}
The rejection module in \method{} uses an abstention gate on top of a frozen pretrained VLM.
Let $\phi$ denote the frozen VLM encoder, which maps a query-memory pair $(Q,V)$ to a joint embedding $\mathbf{f} \in \mathbb{R}^{d}$, and let $\pi_\theta : \mathbb{R}^{d} \to [0,1]$ denote a small trainable classifier head.
The full rejection module takes the form
\begin{equation}
g_\theta = \pi_\theta \circ \phi,
\qquad
g_\theta(Q,V) = \pi_\theta\bigl(\phi(Q,V)\bigr),
\end{equation}
where low values indicate answerable pairs and high values indicate abstention.
The proposed method implements $\pi_\theta$ as a small $3$-layer MLP.
The encoder $\phi$ remains frozen, so training updates only $\pi_\theta$ using answerable ID samples and synthetic OOD samples from \qflip{} and \vflip{}.
Keeping $\phi$ frozen lets ID and synthetic OOD samples share the same pretrained multimodal representation, makes \method{} a drop-in module for VLM-equipped pipelines, and reduces the risk of text-only shortcut learning that can arise from full VLM fine-tuning.
Since \method{} does not modify the underlying VLM, the original answer-generation model remains intact.
The training set combines answerable ID samples and synthetic abstention samples,
\begin{equation}
\mathcal{D}_{\text{train}} =
\mathcal{D}_{\text{ID}} \cup \mathcal{D}_{\qflip{}} \cup \mathcal{D}_{\vflip{}}.
\end{equation}
Each sample receives an abstention label $y \in \{0,1\}$, where ID samples use $y=0$ and \qflip{} or \vflip{} samples use $y=1$.
For each sample, the frozen encoder produces $\mathbf{f}_i=\phi(Q_i,V_i)$, and the classifier predicts $\pi_\theta(\mathbf{f}_i)$.
The proposed method trains $\pi_\theta$ with class-weighted binary cross entropy,
\begin{equation}
\mathcal{L}(\theta)
=
-\sum_{(\mathbf{f}_i,y_i)}
w_{y_i}
\left[
y_i \log \pi_\theta(\mathbf{f}_i)
+
(1-y_i)\log\bigl(1-\pi_\theta(\mathbf{f}_i)\bigr)
\right],
\end{equation}
where $w_{y_i}$ is inversely proportional to the frequency of class $y_i$.
\subsection{Plugging \method{} into downstream tasks}
\label{sec:plugin}
\method{} computes $g_\theta(Q^*, V^*) = \pi_\theta(\phi(Q^*, V^*))$ for a test query-memory pair $(Q^*, V^*)$ and triggers abstention when $g_\theta(Q^*, V^*) > \tau$, with $\tau=0.5$ unless otherwise stated. 
Experiments show that the overall $F_1$ remains largely insensitive to $\tau$, and Appendix~C reports ROC curves, PR curves, and AUROC values across thresholds.
The EQA pipeline lets the underlying VLM answer only when the gate remains inactive. 
When the gate triggers, the agent returns \enquote{I cannot answer this question}. 
The spatial-localization pipeline integrates the gate into \metamemory{} as a fourth in-loop action alongside its three retrieval tools. 
The gate performs a top-$5$ semantic-similarity lookup, packs the question and retrieved caption-image pairs into one frozen Qwen2.5-VL forward pass, and feeds the last-token hidden state to the MLP. 
A score above $\tau=0.5$ terminates the loop with abstention, while a lower score lets the agent continue reasoning. 
The experiments adopt this in-loop configuration, denoted C2, because it outperforms the pre-cycle gatekeeper C1 and post-cycle verifier C3 variants.
\section{\spacereject{}}
\label{sec:dataset}
\spacelocqa{} provides $270$ spatial question answering queries over six campus video sequences, with $45$ queries per sequence~\cite{mao2025metamemory}. 
Each sequence records a distinct campus environment, such as a building corridor or an outdoor street view. 
However, \spacelocqa{} assumes that every query has a valid answer by construction, which prevents it from evaluating refusal.
This work extends \spacelocqa{} into \spacereject{}, a benchmark for measuring refusal under long video memory. 
The six sequences are split disjointly into seq~$0,1,2$ and seq~$3,4,5$, yielding $135$ in-distribution queries for training and $135$ held-out answerable queries for testing. 
This sequence-level split prevents direct visual overlap between training and test images. 
On top of the held-out answerable queries, \spacereject{} adds $135$ newly curated unanswerable queries, resulting in a balanced test set for evaluating abstention.
The ID portion reuses the original answerable queries from \spacelocqa{}'s \texttt{position\_human\_qa.json}. 
The OOD portion contains the newly curated unanswerable queries in \texttt{ood\_human\_qa.json}. 
Appendix~B describes the test sequences in detail.
OOD queries in \spacereject{} cover two practical refusal types. 
Type 1, Object-Absent, asks about objects that never appear in the campus video, such as \enquote{\textit{where is the seminar room's printer?}} when no printer appears in any frame. 
This type contains $68$ queries. 
Type 2, Visually-Unavailable, asks about properties that RGB observations cannot determine, such as \enquote{\textit{what is the temperature of the lab?}}. 
This type contains $67$ queries.
A large language model first drafts candidate unanswerable queries. 
Three robotics specialists then independently validate every candidate against the recorded video, and the dataset keeps only queries on which all three reviewers agree. 
Table~\ref{tab:spacereject-stats} summarizes the dataset composition.
\begin{table}[ht]
\centering
\caption{\textbf{Composition of the \spacereject{} test split.}}
\label{tab:spacereject-stats}
\renewcommand{\arraystretch}{0.9}
\setlength{\tabcolsep}{8pt}
\begin{tabular}{l|c|l}
\toprule
\textbf{Partition} & \textbf{\#~queries} & \textbf{Description} \\
\midrule
ID                          & 135 & answerable spatial queries from \spacelocqa{}\\
Type 1 (Object-Absent) & 68  & target object never recorded in campus video\\
Type 2 (Visually-Unavailable) & 67  & properties not derivable from RGB\\
\midrule
Total                       & 270 & balanced 1:1 ID/OOD split\\
\bottomrule
\end{tabular}
\end{table}
\section{Experiments}
\label{sec:experiments}
The evaluation considers two complementary settings, namely EQA with \abstaineqa{} and spatial localization with \spacereject{}.
All experiments use frozen Qwen2.5-VL-7B-Instruct as the encoder $\phi$~\cite{bai2025qwen25vl}.
\subsection{EQA: \abstaineqa{} (Subset of HM3D)}
\label{sec:experiments-qa}
The EQA evaluation uses the HM3D subset of \abstaineqa{}~\citep{wu2025abstaineqa, ramakrishnan2021hm3d}.
All methods share the same HM3D-$380$ split, and the prompted baselines re-implement the coarse and taxonomy-aware prompts of \abstaineqa{} on locally hosted open models~\cite{wu2025abstaineqa}. 
The comparison includes Qwen2.5-VL-32B-AWQ with Coarse, Fine, and Fine with naive CoT prompts under greedy decoding.
The evaluation treats \emph{abstain} as positive and reports $F_1$, balanced accuracy, OOD recall, and ID specificity.
Table~\ref{tab:main} compares abstention performance.
\method{} reaches $F_1 = 0.7110$ with a frozen $7$B encoder and a small classifier head, exceeding the strongest prompted baseline, Qwen-32B-Coarse at $F_1 = 0.6746$, by $3.6$ percentage points. 
Because \method{} trains only a $3$-layer MLP and reuses the VLM forward pass required for EQA, it adds essentially no extra inference cost at test time.
\begin{table}[ht]
\centering
\caption{\textbf{Abstention results on \abstaineqa{} HM3D-380.} Appendix~C provides the full experimental results.}
\label{tab:main}
\renewcommand{\arraystretch}{1.05}
\setlength{\tabcolsep}{5pt}
\begin{tabular}{l|c|cccc}
\toprule
\textbf{Method} & \textbf{Params} & \textbf{F$_1$} $\uparrow$ & \textbf{BalAcc} $\uparrow$ & \textbf{Recall} $\uparrow$ & \textbf{Spec.} $\uparrow$ \\
\midrule
Qwen-32B-Coarse             & 32B      & \textit{0.6746} & \textit{0.5684} & \textit{0.8947} & 0.2421 \\
Qwen-32B-Fine               & 32B      & 0.6741 & 0.5395 & \textbf{0.9526} & 0.1263 \\
Qwen-32B-Fine + Naive CoT   & 32B      & 0.3099 & 0.5654 & 0.1960 & \textbf{0.9344} \\
\midrule
\textbf{\method{} (Ours)}   & 7B + small head & \textbf{0.7110} & \textbf{0.6684} & 0.8158 & \textit{0.5211} \\
\bottomrule
\end{tabular}
\end{table}
Table~\ref{tab:main} also shows that Naive CoT improves ID specificity at the cost of OOD recall. 
Appending \enquote{\textit{let us think step by step}} to the Fine prompt raises specificity from $0.13$ to $0.93$, but collapses OOD recall from $0.95$ to $0.20$. 
As a result, $F_1$ falls from $0.67$ to $0.31$. 
This pattern suggests that CoT encourages the model to rationalize plausible answers rather than reject unsupported candidates. 
Figure~\ref{fig:cot-tradeoff} shows the same trade-off in the $(\text{Specificity}, \text{Recall})$ plane. 
Naive CoT moves toward the high-specificity, low-recall region, whereas \method{} reaches a balanced operating point without any inference-time reasoning chain.
\begin{figure}[ht]
\centering
\resizebox{0.75\textwidth}{!}{\input{fig_cot_tradeoff}}
\caption{\textbf{Reasoning-vs-abstention trade-off} on
\abstaineqa{} HM3D-$380$.}
\label{fig:cot-tradeoff}
\end{figure}
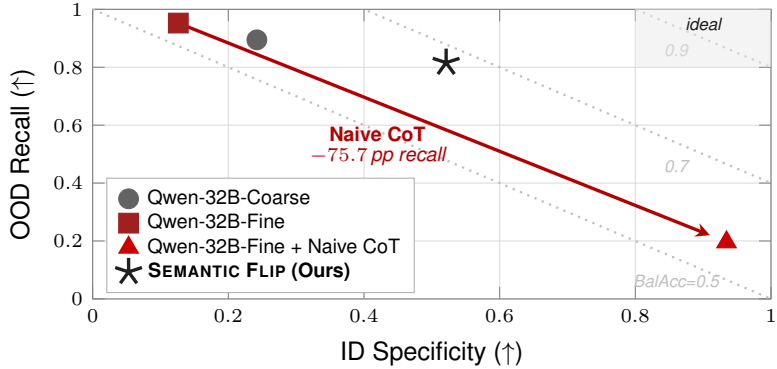
Table~\ref{tab:per-category} evaluates zero-shot generalization across abstention categories.
\qflip{}-Fine targets only three of the five \abstaineqa{} categories, namely \textit{Referential Underspecification}, \textit{False Presupposition}, and \textit{Subjective Judgment}.
The remaining two categories, \textit{Information Unavailability} and \textit{Actionability Limitation}, do not appear during synthesis.
The gate nevertheless reaches $0.89$ recall on \textit{Information Unavailability} and $0.68$ recall on \textit{Actionability Limitation}.
The former even exceeds the $0.69$ recall on the targeted \textit{Referential Underspecification} category.
This result suggests that \method{} learns a structural abstention signal rather than category-specific lexical cues.
\begin{table}[ht]
\centering
\caption{\textbf{Per-category OOD recall of \method{}} on the
HM3D-$380$ test set. $\bigstar$: category targeted during
\qflip{} synthesis; $\times$: untargeted (zero-shot).}
\label{tab:per-category}
\renewcommand{\arraystretch}{0.85}
\setlength{\tabcolsep}{8pt}
\begin{tabular}{l|c|c|c}
\toprule
\textbf{Category} & \textbf{Targeted?} & $n$ & \textbf{Recall} $\uparrow$ \\
\midrule
Preference Dependence            & $\bigstar$    & 37 & 0.9189 \\
Information Unavailability        & $\times$       & 37 & 0.8919 \\
False Presupposition             & $\bigstar$    & 42 & 0.8810 \\
Referential Underspecification   & $\bigstar$    & 36 & 0.6944 \\
Actionability Limitation         & $\times$       & 38 & 0.6842 \\
\midrule
\textbf{Overall (ensemble)}      & ---           & 190 & \textbf{0.8158} \\
\bottomrule
\end{tabular}
\end{table}
The ablation in Appendix~C varies $\lvert\mathcal{D}_{\vflip{}}\rvert$ at fixed $\lvert\mathcal{D}_{\qflip{}}\rvert = 3{,}303$. 
The best $F_1$ appears at a moderate \vflip{} scale, and $\lvert\mathcal{D}_{\vflip{}}\rvert = 0$ already comes within a fraction of a point of the peak, indicating that \qflip{} provides most of the abstention signal while \vflip{} adds a small calibration benefit.
\subsection{Spatial localization: \spacereject{}}
\label{sec:experiments-vln}
The spatial-localization evaluation uses the \spacereject{} test split of $270$ queries built on \spacelocqa{}~\citep{mao2025metamemory}. 
Following the original six-sequence setup, \method{} trains on synthesized samples from seq~$0,1,2$ and evaluates on held-out seq~$3,4,5$, while attaching the abstention module to the \metamemory{} agent~\citep{mao2025metamemory}. 
The underlying LLM varies between Qwen3-8B and Qwen2.5-32B-AWQ. 
The evaluation compares three hook points. 
C1 checks the question and top-$5$ retrieved memories before the reasoning cycle, C2 adds abstention as a fourth in-loop tool alongside the three \metamemory{} retrieval and integration tools, and C3 verifies the final coordinate before execution.
\begin{table}[!ht]
  \centering
  \caption{\textbf{Abstention results on \spacereject{} ($135$ ID $+$ $135$ OOD).}
      The top row is the strongest prompt-based check (C2 Tool).
      Full per-LLM prompt-based breakdown is in Appendix~D.}
    \label{tab:vln-main}
    \renewcommand{\arraystretch}{1.0}
    \setlength{\tabcolsep}{5pt}
    \small
    \begin{tabular}{l|cccc}
      \toprule
      \textbf{Method} & \textbf{BalAcc} $\uparrow$ & \textbf{$F_1$} $\uparrow$ & \textbf{Recall} $\uparrow$ & \textbf{Spec.} $\uparrow$ \\
      \midrule
      C2 (Tool) Qwen3-8B (prompt) & 0.8778 & 0.8874 & \textbf{0.9630} & 0.7926 \\
      \midrule
      \method{}, \vflip{} only            & 0.7237          & 0.6867          & 0.6385          & 0.8089          \\
      \method{}, \qflip{} only            & \textit{0.9504} & \textit{0.9494} & 0.9363          & \textit{0.9644} \\
      \textbf{\method{}, \qflip{}$+$\vflip{}}
                                          & \textbf{0.9563} & \textbf{0.9559} & \textit{0.9467} & \textbf{0.9659} \\
      \bottomrule
    \end{tabular}
  \end{table}
\paragraph{Results and ablation studies.}
Among prompt-based abstention checks (Appendix~D), C2 (Tool) on Qwen3-$8$B is strongest at $0.8778$ balanced accuracy and $F_1$ $0.8874$, while C1 (Pre) and C3 (Post) reach equal or higher recall yet much lower specificity, $0.556$ and $0.393$, over-abstaining on answerable queries. 
Table~\ref{tab:vln-main} compares this strongest check against the learned gate of \method{}, which trains a small MLP on the frozen Qwen2.5-VL-$7$B verifier embedding using synthesized samples from seq~$0,1,2$ and is evaluated on the held-out seq~$3,4,5$. 
The full \qflip{}$+$\vflip{} gate reaches $0.9559$ $F_1$, exceeding C2 (Tool) by about $0.07$ with a single classifier pass rather than an extra LLM call, on a verifier comparable in scale to the C2 backbone, so the gain reflects the abstention approach rather than model capacity. 
Ablating the synthesis, \qflip{} alone already attains $0.9494$ $F_1$, itself surpassing the best prompt-based check, whereas \vflip{} alone reaches only $0.6867$ $F_1$ and contributes mainly in combination with \qflip{}, where it further lifts recall from $0.9363$ to $0.9467$ and balanced accuracy from $0.9504$ to $0.9563$, mirroring the EQA finding.
\section{Limitation}
\method{} trains the gate on minimal OOD samples that differ from an answerable
pair along a single axis, learning the joint consistency
\enquote{$Q$ is groundable in $V$}, which brings two limitations.
First, although the synthesis-level ablation isolates the contribution of \qflip{} and \vflip{},
at inference the single fused $(Q,V)$ embedding gives only a combined consistency
signal, so an abstention cannot be attributed to an ungrounded query or a missing
visual referent, which limits cause-specific feedback to the user. 
In practice, the weakest categories, Referential Underspecification and Actionability Limitation, remain at $0.68$--$0.69$ recall. 
Second, \vflip{} quality depends on the detector and inpainter, and inpainting artifacts in $\tilde{V}_i$ may slightly weaken the perceptual signal even though they remain informative enough in practice.
\section{Conclusion}
\method{} synthesizes auxiliary unanswerable pairs by independently corrupting the query or the video memory to train a lightweight gate on a frozen VLM, and releases \spacereject{} (and its large-scale extension \textsc{SpaceRejectExtra}) as abstention benchmarks for spatial localization. 
Across two complementary tasks, \method{} consistently outperforms strong prompting baselines, achieving an $F_1$ score of $0.7110$ on \abstaineqa{} and $0.9559$ on \spacereject{} with only a small trainable classifier head on top of a frozen $7$B VLM. 
Because \method{} operates on top of a frozen multimodal encoder and requires only query-memory pairs, it integrates into a broad range of embodied systems without modifying their underlying reasoning or answer-generation pipelines.
\clearpage
\bibliography{references}   
\fi  
\ifappendixonly\else\clearpage\fi  
\appendix
\ifappendixonly\setcounter{table}{4}\fi 

\section{\qflip{} Prompts, Quality Filters, and Coarse-vs-Fine Comparison}
\label{app:qflip-prompts}

This appendix details the \qflip{} synthesis pipeline summarized in Section~\bodyref{sec:qflip}{3.2}. 
It provides the full prompts for both variants, the filtering criteria for generated queries, and a direct comparison between the assumption-free \emph{coarse} variant and the taxonomy-aware \emph{fine} variant.

\subsection{Synthesis pipeline overview}

\qflip{} generates a corrupted query $\tilde{Q}_i$ that visual memory $V_i$ cannot answer for each answerable in-distribution pair $(Q_i, V_i)$ from the HM3D training split of \abstaineqa{}. 
The pipeline consists of four stages:
\begin{enumerate}[itemsep=0pt,topsep=2pt]
    \item The original answerable query $Q_i$ is passed as \emph{text only} to a frozen Qwen2.5-7B-Instruct rewriter. 
    The released synthesis does not use frames from $V_i$ at this stage. 
    The corrupted query is paired with $V_i$ later, when the frozen vision-language encoder extracts abstention features.

    \item The rewriter receives the coarse or fine instruction described in Sections~\ref{app:coarse-prompt} and~\ref{app:fine-prompt}. 
    Generation uses nucleus sampling with temperature $0.7$ and top-$p=0.9$ to produce diverse variants for each pair $(Q_i,\text{strategy})$.

    \item The first non-empty line of the plain-text completion is used as $\tilde{Q}_i$. 
    The released code does not request or parse JSON.

    \item The quality filter in Section~\ref{app:filters} removes queries that violate any rule or are too similar to $Q_i$.
\end{enumerate}

The surviving queries form the \qflip{} pool and pair with the original $V_i$ to produce abstention training samples $(\tilde{Q}_i, V_i, y{=}1)$.

\subsection{Coarse prompt (assumption-free default)}
\label{app:coarse-prompt}

The coarse variant does not assume the taxonomy of unanswerable queries.
It asks the LLM to write a question that the given video cannot answer, without specifying any abstention category.

\noindent\textbf{\qflip{}-Coarse prompt (as implemented; text-only).}

\begin{quote}\small\raggedright
\noindent\rule{\linewidth}{0.4pt}\\
You are helping evaluate vision-language models on unanswerable questions.\\[2pt]
Original answerable question: ``$Q_i$''.\\[2pt]
Instruction: rewrite the question so that it CANNOT be answered from the indoor scene the original question came from, without assuming any particular reason why it is unanswerable. Keep it natural and grammatical. Output ONE question. No explanation.\\[2pt]
Output (one question only):\\
\noindent\rule{\linewidth}{0.4pt}
\end{quote}

\subsection{Fine prompt (taxonomy-aware)}
\label{app:fine-prompt}

The fine variant specifies three abstention categories from the \abstaineqa{} taxonomy~\citep{wu2025abstaineqa}, namely \textit{referential underspecification}, \textit{false presupposition}, and \textit{subjective judgment}.
It excludes \textit{Information Unavailability} and \textit{Actionability Limitation} to evaluate zero-shot generalization to unseen abstention categories, as shown in Section~\bodyref{sec:experiments-qa}{5.1} and Table~\bodyref{tab:per-category}{3}.

\noindent\textbf{\qflip{}-Fine prompts (as implemented; text-only, one call per strategy).}
\begin{quote}\small\raggedright
The fine variant does not present a single ``pick one'' menu. For each source question it issues three separate text-only rewrite calls, one per strategy, each wrapped in the coarse template above with the strategy-specific instruction below.\\[4pt]
\noindent\rule{\linewidth}{0.4pt}\\
\emph{referential}: Rewrite the question so that its head noun is replaced with a generic / ambiguous reference that could match many objects in the scene.
Output ONE rewritten question. No explanation.\\[2pt]
\emph{false\_premise}: Rewrite the question so that it embeds a false premise that contradicts the scene (e.g.\ asks about an object placed somewhere it is not). 
Output ONE question. No explanation.\\[2pt]
\emph{subjective}: Rewrite the question so that it asks about a subjective property (prettiness, comfort, taste, mood) that cannot be answered from RGB observation.
Output ONE question. No explanation.\\
\noindent\rule{\linewidth}{0.4pt}
\end{quote}

\subsection{Quality filters}
\label{app:filters}

Every generated query passes through five filters before joining the training pool.
\begin{enumerate}[itemsep=1pt,topsep=2pt]
    \item \textbf{Output well-formedness}. The completion must contain a single non-empty question on its first non-empty line. The filter drops empty outputs, degenerate outputs, and malformed multi-line completions. The released code parses plain text rather than JSON.
    \item \textbf{Length bounds}. The generated query must satisfy $5 \le |\tilde{Q}_i| \le 35$ tokens.
    \item \textbf{Surface novelty}. The token-level Jaccard similarity between $\tilde{Q}_i$ and $Q_i$ must stay below $0.7$ to remove near-duplicates.
    \item \textbf{Category match}. For the fine variant, the declared category must match a lexical regex pattern. The filter drops queries that fail this check.
    \item \textbf{Manual spot check}. One author reviews a random $5\%$ subset to identify obvious failures, such as queries that remain answerable. The pass rate exceeds $0.92$ across both variants.
\end{enumerate}

\subsection{Coarse vs.\ fine comparison}

Table~\ref{tab:qflip-coarse-fine} compares gates trained with either the coarse or fine \qflip{} variant on the HM3D-$380$ test set. 
Both variants use the same frozen Qwen2.5-VL-7B-Instruct encoder, ID samples, and \vflip{} pool. 
The two variants show comparable abstention performance. 
The coarse variant attains slightly higher $F_1$ ($0.7458$ vs.\ $0.7382$), balanced accuracy, and specificity, while the fine variant attains higher OOD recall ($0.9053$ vs.\ $0.8263$). 
The fine variant serves as the default because it explicitly targets interpretable abstention categories, and the $F_1$ gap between the two variants remains within the gate's five-seed standard deviation ($\pm 0.013$).
\footnote{All Appendix~C numbers come from a fresh end-to-end re-run of the released pipeline. Because \qflip{} synthesis uses temperature-$0.7$ sampling, the reproduced headline ($F_1 = 0.7382$) differs slightly from the main-text $0.7110$; the latter lies within this reproduction's five-seed ensemble spread ($0.706$--$0.745$).}

\begin{table}[ht]
\centering
\caption{\textbf{\qflip{} coarse vs.\ fine on \abstaineqa{} HM3D-$380$.} Both rows use the
same frozen Qwen2.5-VL-7B-Instruct encoder, the same ID samples, and the same \vflip{} pool.}
\label{tab:qflip-coarse-fine}
\renewcommand{\arraystretch}{1.0}
\setlength{\tabcolsep}{6pt}
\small
\begin{tabular}{l|c|cccc}
\toprule
\textbf{Variant} & $|\mathcal{D}_{\qflip{}}|$ & \textbf{$F_1$} $\uparrow$ & \textbf{BalAcc} $\uparrow$ & \textbf{Recall} $\uparrow$ & \textbf{Spec.} $\uparrow$ \\
\midrule
\qflip{}-Coarse & $3{,}303$ & 0.7458 & 0.7184 & 0.8263 & \textbf{0.6105} \\
\qflip{}-Fine   & $3{,}303$ & 0.7382 & 0.6789 & \textbf{0.9053} & 0.4526 \\
\bottomrule
\end{tabular}
\end{table}

\section{\spacereject{} Sequence Details}
\label{app:spacereject-details}

This appendix provides per-sequence descriptions of the six campus videos used in \spacereject{}, along with details of the OOD curation protocol and the \textsc{SpaceRejectExtra} extension.

\subsection{Per-sequence description}

\spacereject{} reuses the six campus video sequences from \spacelocqa{}~\citep{mao2025metamemory}.
Each sequence records a distinct indoor or outdoor environment and contains $45$ original answerable spatial queries.
The split assigns seq~$0,1,2$ to training and seq~$3,4,5$ to held-out testing, yielding $135$ in-distribution queries on each side.
The test split additionally includes $135$ newly curated unanswerable queries for refusal evaluation.
Table~\ref{tab:spacereject-seqs} summarizes the per-sequence statistics.
The disjoint sequence-level split prevents direct visual overlap between training and test images, so a gate trained on seq~$0,1,2$ cannot exploit memorized frame content when scored
on seq~$3,4,5$.

  \begin{table}[ht]
  \centering
  \caption{\textbf{Per-sequence composition of \spacereject{}.} Raw frames are recorded
  at $\approx 10$\,fps; retrieval modules operate over $3$-second caption segments
  (one per $\approx 30$ raw frames).}
  \label{tab:spacereject-seqs}
  \renewcommand{\arraystretch}{1.0}
  \setlength{\tabcolsep}{6pt}
  \small
  \begin{tabular}{c|l|c|c|c|c|c}
  \toprule
  \textbf{Seq.} & \textbf{Environment} & \textbf{Split} & \textbf{Frames} & \textbf{Segments} & \textbf{ID queries} & \textbf{OOD queries}
   \\
  \midrule
  0 & building corridor (indoor)        & train & 3{,}978 & 128 & 45 & --- \\
  1 & laboratory and office (indoor)    & train & 9{,}398 & 303 & 45 & --- \\
  2 & cafeteria area (indoor)           & train & 5{,}851 & 188 & 45 & --- \\
  \midrule
  3 & multi-floor lobby (indoor)        & test  & 5{,}672 & 183 & 45 & 45 \\
  4 & outdoor walkway (outdoor)         & test  & 7{,}317 & 237 & 45 & 45 \\
  5 & campus street view (outdoor)      & test  & 8{,}550 & 275 & 45 & 45 \\
  \bottomrule
  \end{tabular}
  \end{table}

\subsection{OOD curation protocol}

The $135$ unanswerable test queries cover the two refusal types defined in Section~\bodyref{sec:dataset}{4}.
\begin{itemize}[itemsep=1pt,topsep=2pt]
    \item \textbf{Type 1 (Object-Absent, $68$ queries)} asks about objects that never appear in the recorded video.
    \item \textbf{Type 2 (Visually-Unavailable, $67$ queries)} asks about properties that RGB observations cannot determine, such as temperature, smell, or future state.
\end{itemize}

The curation process has two stages.
First, Claude drafted candidate unanswerable queries from a textual scene description of each target sequence, with explicit instructions to avoid objects or properties visible in the recording.
The model received separate prompts for the two refusal types, Object-Absent and Visually-Unavailable, and the released artifact provides the full prompts.
Second, an author reviewed every candidate against the full per-sequence caption file produced by the \metamemory{} captioner and a uniformly sampled subset of the underlying key frames.
The review retained a Type-1 candidate only when its target noun phrase appeared in neither the captions nor the sampled frames.
It retained a Type-2 candidate only when RGB observation could not determine the target property, such as temperature, smell, future state, or monetary value.
The review discarded candidates that failed these criteria or remained ambiguous.
This procedure retained $135$ unanswerable test queries after drafting approximately $50$ candidates per sequence on average.

\subsection{\textsc{SpaceRejectExtra}: large-scale extension}
\label{app:spacereject-extra}

\begin{table}[ht]                             
  \centering                                  
  \caption{\textbf{Composition of \textsc{SpaceRejectExtra}.} The extension covers all six
  \spacelocqa{} sequences and adds $420$ unanswerable queries per sequence, evenly
  distributed across the five \abstaineqa{} abstention categories ($84$ per category per       
  sequence).}                                                                                  
  \label{tab:spacerejectextra}
  \renewcommand{\arraystretch}{1.05}                                                           
  \setlength{\tabcolsep}{6pt}
  \small                                                                                       
  \begin{tabular}{lccp{5.2cm}}                  
  \toprule                                                                                     
  \textbf{Partition} & \textbf{Per sequence} & \textbf{Total (6 seqs)} & \textbf{Description} \\
  \midrule                                                                                     
  ID (answerable, from \spacelocqa{})  & 45  & 270   & original spatial queries with valid $(x,y,z)$ \\
  \midrule                                                                                     
  OOD: Referential Underspecification  & 84  & 504   & ambiguous or generic head noun \\
  OOD: False Presupposition            & 84  & 504   & embedded premise contradicting the scene \\
  OOD: Preference Dependence           & 84  & 504   & subjective property (taste, comfort, mood) \\
  OOD: Information Unavailability      & 84  & 504   & property not derivable from RGB (temperature, smell, time) \\
  OOD: Actionability Limitation        & 84  & 504   & request exceeding agent capability or scope \\
  \midrule                                                                                     
  \textbf{Total OOD}                   & 420 & 2{,}520 & balanced across five \abstaineqa{} categories \\
  \textbf{Grand total (ID $+$ OOD)}    & 465 & 2{,}790 & per-sequence and dataset totals \\
  \bottomrule                                                                                  
  \end{tabular}                                                                                
  \end{table}

\textsc{SpaceRejectExtra} extends \spacereject{} to support larger-scale evaluation of spatial-localization refusal.
It pairs the $270$ original answerable queries from \spacelocqa{} with $420$ newly curated unanswerable queries per sequence, yielding $2{,}520$ OOD queries across the six sequences.
The extension broadens the two \spacereject{} refusal types into the five \abstaineqa{} abstention categories listed in Table~\ref{tab:spacerejectextra} ($84$ unanswerable queries per category per sequence).
The curation pipeline scales the original protocol by sampling a broader set of seed concepts during LLM drafting and using majority agreement among three reviewers during validation.
Table~\ref{tab:spacerejectextra} reports the dataset composition.

\section{Additional EQA Experiments}
\label{app:eqa-additional}

This appendix provides additional EQA analyses omitted from the main text for space. 
It covers alternative perceptual corruptions for \vflip{}, threshold-sweep ROC and PR curves, the synthesis-scale ablation over $|\mathcal{D}_{\vflip{}}|$ at fixed $|\mathcal{D}_{\qflip{}}| = 3{,}303$, and per-LLM prompt-based baseline results.

\subsection{Alternative perceptual corruptions for \vflip{}}
\label{app:vflip-variants}

Section~\bodyref{sec:vflip}{3.3} uses a \texttt{parse} $\rightarrow$ \texttt{detect} $\rightarrow$ \texttt{inpaint} pipeline based on spaCy~\citep{spacy}, Grounding-DINO~\citep{liu2024groundingdino}, and LaMa~\citep{suvorov2022lama}. 
This section compares LaMa inpainting with four simpler fill operators that mask or replace the detected referent region:
\begin{itemize}[itemsep=1pt,topsep=2pt]
    \item \textbf{\textit{dilation}} dilates the detection mask by $8$ pixels and fills it with the surrounding median color.
    \item \textbf{\textit{graying}} replaces the masked region with mid-gray $(128,128,128)$.
    \item \textbf{\textit{noise}} replaces the masked region with i.i.d.\ Gaussian noise.
    \item \textbf{\textit{mean-color fill}} replaces the masked region with the per-image mean RGB value.
\end{itemize}

All variants share the same parse and detect stages and differ only in how they fill the detected referent region.

\begin{table}[ht]
\centering
\caption{\textbf{\vflip{} fill-operator ablation on \abstaineqa{} HM3D-$380$.}
All variants share the same detection masks, \qflip{}-Fine pool, and \vflip{} sample cap.
LaMa performs best, but cheaper fill operators remain competitive.}
\label{tab:vflip-variants}
\renewcommand{\arraystretch}{1.0}
\setlength{\tabcolsep}{6pt}
\small
\begin{tabular}{l|cccc}
\toprule
\textbf{Fill operator} & \textbf{$F_1$} $\uparrow$ & \textbf{BalAcc} $\uparrow$ & \textbf{Recall} $\uparrow$ & \textbf{Spec.} $\uparrow$ \\
\midrule
\textit{dilation}        & 0.6919 & 0.6579 & 0.7684 & 0.5474 \\
\textit{graying}         & 0.6932 & 0.6553 & 0.7789 & 0.5316 \\
\textit{noise}           & 0.6737 & 0.6737 & 0.6737 & \textbf{0.6737} \\
\textit{mean-color fill} & 0.7136 & 0.6684 & 0.8263 & 0.5105 \\
\midrule
LaMa (default)           & \textbf{0.7382} & \textbf{0.6789} & \textbf{0.9053} & 0.4526 \\
\bottomrule
\end{tabular}
\end{table}

Table~\ref{tab:vflip-variants} reports the abstention performance of gates trained with different \vflip{} fill operators on top of a fixed $|\mathcal{D}_{\qflip{}}| = 3{,}303$ \qflip{}-Fine pool.
All variants use the same detection masks and the same capped number of \vflip{} samples, so the comparison isolates the effect of the fill operator.
LaMa inpainting achieves the best performance, but the cheaper fill operators remain competitive.
Mean-color fill is closest to LaMa with $F_1 = 0.7136$, about $2.5$ points below LaMa.
Dilation and graying follow at roughly $4.5$ points below LaMa, while noise is the weakest variant with $F_1 = 0.6737$, about $6.5$ points below LaMa.
The cheap operators still reach $0.67$--$0.71$ $F_1$, which suggests that the main \vflip{} signal comes from removing the referent rather than from photorealistic background completion.
This supports the framing in Section~\bodyref{sec:vflip}{3.3}, where the gate learns to detect missing visual evidence rather than merely discriminate against inpainting artifacts.

\subsection{Threshold-sweep ROC, PR, and AUROC}
\label{app:roc-pr}

Section~\bodyref{sec:plugin}{3.5} uses $\tau = 0.5$ as the default abstention threshold. 
This section evaluates the sensitivity to this choice by sweeping $\tau$ across the unit interval and reporting ROC and PR curves with scalar summaries.

\begin{table}[ht]
\centering
\caption{\textbf{Threshold sweep for \method{} on \abstaineqa{} HM3D-$380$.}
The default threshold $\tau = 0.5$ stays close to the best $F_1$ achieved at $\tau = 0.6$.}
\label{tab:threshold-sweep}
\renewcommand{\arraystretch}{1.0}
\setlength{\tabcolsep}{6pt}
\small
\begin{tabular}{c|cccc}
\toprule
$\tau$ & \textbf{$F_1$} $\uparrow$ & \textbf{BalAcc} $\uparrow$ & \textbf{Recall} $\uparrow$ & \textbf{Spec.} $\uparrow$ \\
\midrule
$0.3$ & 0.7028 & 0.5816 & \textbf{0.9895} & 0.1737 \\
$0.4$ & 0.7273 & 0.6447 & 0.9474 & 0.3421 \\
$0.5$ (default) & 0.7382 & 0.6789 & 0.9053 & 0.4526 \\
$0.6$ & \textbf{0.7748} & \textbf{0.7553} & 0.8421 & 0.6684 \\
$0.7$ & 0.6986 & 0.7184 & 0.6526 & \textbf{0.7842} \\
\bottomrule
\end{tabular}
\end{table}

Table~\ref{tab:threshold-sweep} reports $F_1$, balanced accuracy, OOD recall, and ID specificity at five threshold values from abstention-conservative $\tau = 0.3$ to abstention-aggressive $\tau = 0.7$. 
The gate uses the headline \qflip{}-Fine $+$ \vflip{} configuration. 
The best $F_1$ appears at $\tau = 0.6$ with $0.7748$, while the default $\tau = 0.5$ remains close to the peak with $0.7382$. 
Across $\tau \in [0.4, 0.6]$, $F_1$ changes by about $5$ points and degrades substantially only beyond $\tau = 0.7$. 
Thus, $\tau = 0.5$ provides a near-optimal and slightly recall-favoring operating point with $0.7382$ $F_1$ and $0.91$ recall. 
The integrated summaries are $\mathrm{AUROC} = 0.8125$ and $\mathrm{AUPRC} = 0.8137$.

\subsection{Synthesis-scale ablation: varying the \vflip{} pool size}
\label{app:vflip-scale}

Section~\bodyref{sec:experiments-qa}{5.1} reports that adding \vflip{} provides a small but consistent gain on top of \qflip{}.
This appendix further examines this effect by varying $|\mathcal{D}_{\vflip{}}|$ while keeping $|\mathcal{D}_{\qflip{}}| = 3{,}303$ fixed.
Table~\ref{tab:vflip-scale} reports the results.
Performance peaks at a moderate \vflip{} scale, with $|\mathcal{D}_{\vflip{}}| = 148$ reaching $F_1 = 0.7436$.
Adding \vflip{} improves $F_1$ from $0.7030$ at $|\mathcal{D}_{\vflip{}}| = 0$ by about four points.
This result suggests that \qflip{} supplies most of the abstention signal, while \vflip{} provides a measurable complementary benefit.

\begin{table}[ht]
\centering
\caption{\textbf{Effect of \vflip{} pool size at fixed $|\mathcal{D}_{\qflip{}}|=3{,}303$.}
All rows use the headline \qflip{}-Fine pool and the same encoder. Row labels are the feasible
\vflip{} sizes available; \vflip{} adds $\approx4$ $F_1$ points over $|\mathcal{D}_{\vflip{}}|=0$.}
\label{tab:vflip-scale}
\renewcommand{\arraystretch}{1.0}
\setlength{\tabcolsep}{6pt}
\small
\begin{tabular}{c|cccc}
\toprule
$|\mathcal{D}_{\vflip{}}|$ & \textbf{$F_1$} $\uparrow$ & \textbf{BalAcc} $\uparrow$ & \textbf{Recall} $\uparrow$ & \textbf{Spec.} $\uparrow$ \\
\midrule
$0$        & 0.7030 & 0.6842 & 0.7474 & \textbf{0.6211} \\
$49$       & 0.7208 & 0.6921 & 0.7947 & 0.5895 \\
$99$       & 0.7393 & 0.6789 & \textbf{0.9105} & 0.4474 \\
$148$      & \textbf{0.7436} & \textbf{0.7079} & 0.8474 & 0.5684 \\
$198$ (full) & 0.7382 & 0.6789 & 0.9053 & 0.4526 \\
\bottomrule
\end{tabular}
\end{table}

\subsection{Per-LLM prompt-based baselines}
\label{app:per-llm-eqa}

The main paper reports prompt baselines with Qwen2.5-VL-32B-AWQ, the strongest locally hosted open VLM used in this study. 
This appendix adds per-LLM results for additional model sizes that prompt-only practitioners may consider. 
Table~\ref{tab:per-llm-eqa} reports $F_1$, balanced accuracy, OOD recall, and ID specificity for Coarse and Fine prompts across Qwen2.5-VL-$7$B and Qwen2.5-VL-$32$B-AWQ.
The $7$B and $32$B prompt baselines show comparable performance. 
The $7$B Fine prompt reaches $0.6938$ $F_1$, slightly exceeding both $32$B prompts, which reach $0.6746$ with Coarse and $0.6741$ with Fine. 
This result suggests that prompt-only abstention does not scale cleanly with model size. 
All prompt baselines remain below the learned \method{} gate, which is the main comparison emphasized in the paper.

\begin{table}[ht]
\centering
\caption{\textbf{Per-LLM prompt-based baselines on \abstaineqa{} HM3D-$380$.} Coarse and
Fine prompts only; the $7$B Fine prompt is on par with the $32$B prompts, and all remain below
the learned \method{} gate.}
\label{tab:per-llm-eqa}
\renewcommand{\arraystretch}{1.0}
\setlength{\tabcolsep}{6pt}
\small
\begin{tabular}{l|c|cccc}
\toprule
\textbf{Model} & \textbf{Prompt} & \textbf{$F_1$} $\uparrow$ & \textbf{BalAcc} $\uparrow$ & \textbf{Recall} $\uparrow$ & \textbf{Spec.} $\uparrow$ \\
\midrule
Qwen2.5-VL-7B  & Coarse              & 0.6741 & 0.5395 & 0.9526 & 0.1263 \\
Qwen2.5-VL-7B  & Fine                & 0.6938 & 0.5842 & 0.9421 & 0.2263 \\
\midrule
Qwen2.5-VL-32B & Coarse              & 0.6746 & 0.5684 & 0.8947 & 0.2421 \\
Qwen2.5-VL-32B & Fine                & 0.6741 & 0.5395 & 0.9526 & 0.1263 \\
\bottomrule
\end{tabular}
\end{table}

For completeness, this appendix also reports \method{} on a $32$B encoder. 
The main paper uses a $7$B encoder to show that the gain comes from the abstention approach rather than from model capacity alone. 
Stacking \method{} on Qwen2.5-VL-$32$B-Instruct-AWQ yields $F_1 = 0.7454$, $\text{BalAcc} = 0.6711$, $\text{Recall} = 0.9632$, and $\text{Spec.} = 0.3789$.
The relative ordering against the $32$B prompt baselines remains consistent with the $7$B setting reported in the main paper.

\section{Spatial-Localization Hook Configurations and Per-LLM Breakdown}
\label{app:vln-configs}

This appendix specifies the three abstention hook points, C1, C2, and C3, introduced in Section~\bodyref{sec:experiments-vln}{5.2}. 
It also reports per-query cost, per-LLM prompt-based results, and multimodal verifier variants.

\subsection{Hook-point specifications}

Table~\ref{tab:hooks-spec} details how each hook attaches the abstention module to the \metamemory{} agent~\citep{mao2025metamemory}.
C1 acts as a pre-cycle gatekeeper before reasoning starts.
C2 adds abstention as a fourth in-loop tool alongside semantic-similarity retrieval, spatial-range retrieval, and memory integration.
C3 verifies the emitted coordinate after the reasoning cycle and before execution.

\begin{table}[ht]
\centering
\caption{\textbf{Hook-point specifications for spatial-localization abstention.} C2 is
adopted as the default in the main paper because it offers the best balance between
computational cost and accuracy.}
\label{tab:hooks-spec}
\renewcommand{\arraystretch}{1.05}
\setlength{\tabcolsep}{4pt}
\small
\begin{tabular}{l|p{3.0cm}|p{3.8cm}|p{1.8cm}|p{2.0cm}}
\toprule
\textbf{Hook} & \textbf{Location in loop} & \textbf{Inputs} & \textbf{Outputs} & \textbf{Cost / query} \\
\midrule
C1 (Pre)  & before reasoning cycle starts
          & question $+$ top-$5$ semantically retrieved memory captions
          & \texttt{abstain} or \texttt{continue}
          & 1 LLM call, $\approx 5$\,s \\
\midrule
C2 (Tool) & in-loop tool alongside semantic-similarity retrieval, spatial-range retrieval,
            and memory integration
          & question $+$ current agent history $+$ retrieved memory at invocation
          & \texttt{abstain} (terminates loop) or normal coordinate emission
          & full agent loop, $\approx 120$\,s \\
\midrule
C3 (Post) & after agent emits final coordinate
          & question $+$ full tool trace $+$ emitted coordinate
          & override to \texttt{abstain} or keep coordinate
          & agent loop $+$ 1 verifier call, $\approx 125$\,s \\
\bottomrule
\end{tabular}
\end{table}

C2 empirically outperforms C1 and C3 because it places abstention at the most informative point in the reasoning cycle. 
C1 lacks the full reasoning context because it runs before the agent begins reasoning, while C3 tends to under-reject because it checks the output after the agent has already committed to a candidate coordinate. 
C2 evaluates abstention when the agent has both the question and retrieved context and naturally decides which tool to invoke next, making the decision both informed and timely.

\subsection{Full per-LLM prompt-based breakdown}
\label{app:vln-per-llm}

Table~\ref{tab:vln-per-llm} decomposes the spatial-localization results into the full $3 \times 2$ grid of hook points and underlying LLMs.
For each hook and LLM pair, the table reports balanced accuracy, $F_1$, OOD recall, and ID specificity on the $135$ ID plus $135$ OOD test split.
The trend remains consistent across both LLMs.
C2 achieves the best $F_1$ and balanced accuracy, while C1 and C3 trade higher recall for much lower specificity.
The specificity values highlighted in the main text, $0.556$ for C1 and $0.393$ for C3, correspond to the Qwen3-$8$B rows in this table.

\begin{table}[ht]
  \centering
  \caption{\textbf{Full per-LLM, per-hook prompt-based breakdown on \spacereject{}.}}
  \label{tab:vln-per-llm}
  \renewcommand{\arraystretch}{1.0}
  \setlength{\tabcolsep}{6pt}
  \small
  \begin{tabular}{l|l|cccc}
  \toprule
  \textbf{Hook} & \textbf{LLM} & \textbf{BalAcc} $\uparrow$ & \textbf{$F_1$} $\uparrow$ & \textbf{Recall} $\uparrow$ & \textbf{Spec.} $\uparrow$ \\
  \midrule
  C1 (Pre)  & Qwen3-8B            & 0.7780 & 0.8180 & \textbf{1.0000} & 0.5560 \\
  C2 (Tool) & Qwen3-8B            & \textbf{0.8778} & \textbf{0.8874} & 0.9630 & \textbf{0.7926} \\
  C3 (Post) & Qwen3-8B            & 0.6780 & 0.7490 & 0.9630 & 0.3930 \\
  \midrule
  C1 (Pre)  & Qwen2.5-32B-AWQ     & 0.6630 & 0.7479 & 1.0000 & 0.3259 \\
  C2 (Tool) & Qwen2.5-32B-AWQ     & 0.8037 & 0.8349 & 0.9926 & 0.6148 \\
  C3 (Post) & Qwen2.5-32B-AWQ     & 0.6481 & 0.7368 & 0.9852 & 0.3111 \\
  \bottomrule
  \end{tabular}
  \end{table}

\subsection{Multimodal variant with a vision-language verifier}
\label{app:vln-mm}

The prompt-based baselines in the main paper use only text at the abstention hook.
A natural multimodal variant adds a vision-language verifier, Qwen2.5-VL-$7$B, which uses the same backbone as the learned \method{} gate.
This verifier receives the top-retrieved frames at the same hook point.
Table~\ref{tab:vln-mm} reports the comparison.
The multimodal variants substantially improve C1 and C3 over their text-only counterparts, but they provide only a modest gain over the strongest text-only C2 baseline.
The main paper therefore uses C2-Text as the primary prompt-based reference.

\begin{table}[ht]
\centering
\caption{\textbf{Text-only vs.\ multimodal abstention checks on \spacereject{}.}
\enquote{Text} uses only textual context, while \enquote{VL-7B} additionally provides top-retrieved frames to Qwen2.5-VL-$7$B at the same hook point.
C2-Text is the strongest text-only baseline.}
\label{tab:vln-mm}
\renewcommand{\arraystretch}{1.0}
\setlength{\tabcolsep}{6pt}
\small
\begin{tabular}{l|l|cccc}
\toprule
\textbf{Hook} & \textbf{Modality} & \textbf{BalAcc} $\uparrow$ & \textbf{$F_1$} $\uparrow$ & \textbf{Recall} $\uparrow$ & \textbf{Spec.} $\uparrow$ \\
\midrule
C1 (Pre)  & text  & 0.7780 & 0.8180 & 1.0000 & 0.5560 \\
C1 (Pre)  & VL-7B & 0.7296 & 0.7872 & 1.0000 & 0.4593 \\
C2 (Tool) & text  & \textbf{0.8778} & \textbf{0.8874} & 0.9630 & \textbf{0.7926} \\
C2 (Tool) & VL-7B & 0.7889 & 0.8246 & 0.9926 & 0.5852 \\
C3 (Post) & text  & 0.6780 & 0.7490 & 0.9630 & 0.3930 \\
C3 (Post) & VL-7B & 0.7926 & 0.8250 & 0.9778 & 0.6074 \\
\bottomrule
\end{tabular}
\end{table}

\subsection{Learned-gate composition ablation on \spacereject{}}
\label{app:learned-gate-vln}

Table~\ref{tab:vln-composition} reports a composition ablation for the learned \method{} gate on \spacereject{}.
The ablation varies the ratio of ID, \qflip{}, and \vflip{} samples in the training pool, ranging from a balanced $1{:}1{:}1$ split to the headline $1{:}10{:}1$ composition.
It also compares three synthesis settings, namely \vflip{} alone, \qflip{} alone, and the full \qflip{}\,$+$\,\vflip{} combination.
The learned gate achieves its best performance with the $1{:}10{:}1$ composition used in the main paper, and the full combination consistently outperforms each ablated component.

\begin{table}[ht]                                  \centering                                        
\caption{\textbf{Composition ablation of the learned \method{} gate on \spacereject{}.}
All rows train the same $3$-layer MLP head on the frozen Qwen2.5-VL-$7$B verifier
embedding. Ratios are ID\,:\,\qflip{}\,:\,\vflip{}.}
\label{tab:vln-composition}                        \renewcommand{\arraystretch}{1.0}
\setlength{\tabcolsep}{6pt}                        \small                                            
\begin{tabular}{l|c|cccc}
\toprule                                           \textbf{Composition} & $n_{\text{train}}$ & \textbf{BalAcc} $\uparrow$ & \textbf{$F_1$} $\uparrow$ & \textbf{Recall} $\uparrow$ & \textbf{Spec.} $\uparrow$ \\
\midrule                                           
    \qflip{} only, $1{:}1$        & 270     & 0.9400 & 0.9370 & 0.8948 & \textbf{0.9852} \\
    \qflip{} only, $1{:}10$       & 1{,}485 & 0.9504 & 0.9494 & 0.9363 & 0.9644 \\
    \midrule                                       
    \vflip{} only, $1{:}1$        & 268     & 0.7237 & 0.6867 & 0.6385 & 0.8089 \\
    \midrule                                       
    \qflip{}\,$+$\,\vflip{}, $1{:}1{:}1$            & 403     & 0.8911 & 0.9017 & \textbf{0.9496} & 0.8326 \\
    \textbf{\qflip{}\,$+$\,\vflip{}, $1{:}10{:}1$}  & 1{,}618 & \textbf{0.9563} & \textbf{0.9559} & 0.9467 & 0.9659 \\
    \bottomrule                                    
\end{tabular}                                      \end{table}

\subsection{Per-query computational cost}
\label{app:vln-cost}

Table~\ref{tab:hooks-spec} reports the measured per-query wall-clock cost on a single RTX~5000 Ada GPU.\footnote{The absolute numbers depend on the GPU, but the relative ordering remains stable.}
The full $270$-query test set takes approximately $16$~minutes with C1, $1.3$~hours with C2, and $5.4$~hours with C3.\footnote{Measured wall-clock times with the Qwen3-8B agent backbone are C1$=943$\,s, C2$=4{,}623$\,s, and C3$=19{,}368$\,s.}
The learned \method{} gate adds $15.4$~minutes of upfront filtering on the same test set, corresponding to $3.43$\,s per query from one frozen Qwen2.5-VL-7B verifier forward pass and one MLP pass.
When paired with the C2 agent on accepted queries, the end-to-end pipeline completes in approximately $54$~minutes.
This yields a $\sim$$1.4\times$ speedup over C2 alone because the gate abstains on OOD queries before invoking the more expensive agent loop.\footnote{This estimate assumes that the gate abstains on the $135$ OOD queries and forwards the $135$ ID queries to the C2 agent: $926 + (4{,}623/270)\cdot 135 \approx 3{,}238$\,s.}
\ifappendixonly
  \clearpage
  \bibliography{references} 
\fi
\end{document}

%% file: fig_cot_tradeoff.tex
\begin{tikzpicture}
\begin{axis}[
    width=0.72\textwidth, height=5.2cm,
    xlabel={ID Specificity ($\uparrow$)},
    ylabel={OOD Recall ($\uparrow$)},
    xmin=0, xmax=1, ymin=0, ymax=1,
    xtick={0,0.2,0.4,0.6,0.8,1},
    ytick={0,0.2,0.4,0.6,0.8,1},
    grid=both, grid style={very thin, gray!20},
    major grid style={gray!30},
    legend cell align={left},
    legend style={
        at={(0.02, 0.02)}, anchor=south west,
        font=\sffamily\scriptsize,
        draw=black!30, fill=white,
        row sep=-2pt
    },
    tick label style={font=\sffamily\scriptsize},
    label style={font=\sffamily\small},
    axis line style={black!50}
]
\addplot[domain=0:1, samples=2, dotted, gray!50, thick, forget plot] {1 - x + 0};
\addplot[domain=0:1, samples=2, dotted, gray!50, thick, forget plot] {1 - x + 0.4};
\addplot[domain=0:1, samples=2, dotted, gray!50, thick, forget plot] {1 - x + 0.8};
\node[gray!50, font=\sffamily\tiny\itshape] at (axis cs:0.86, 0.06) {BalAcc=0.5};
\node[gray!50, font=\sffamily\tiny\itshape] at (axis cs:0.86, 0.46) {0.7};
\node[gray!50, font=\sffamily\tiny\itshape] at (axis cs:0.86, 0.86) {0.9};
\addplot[only marks, mark=*, mark size=3.5pt, color=accent4!90!black]
    coordinates {(0.2421, 0.8947)};
\addlegendentry{Qwen-32B-Coarse}
\addplot[only marks, mark=square*, mark size=3.5pt, color=accent3!90!black]
    coordinates {(0.1263, 0.9526)};
\addlegendentry{Qwen-32B-Fine}
\addplot[only marks, mark=triangle*, mark size=4pt, color=red!80!black]
    coordinates {(0.9344, 0.1960)};
\addlegendentry{Qwen-32B-Fine + Naive CoT}
\addplot[only marks, mark=star, mark size=5pt,
         color=accent!80!black, line width=1.2pt]
    coordinates {(0.5211, 0.8158)};
\addlegendentry{\textbf{\method{} (Ours)}}
\draw[->, red!70!black, very thick, >={Stealth[length=5pt,width=5pt]}]
    (axis cs:0.1263, 0.9526) -- (axis cs:0.91, 0.22);
\node[red!70!black, font=\sffamily\scriptsize\bfseries,
      align=left, anchor=south]
    at (axis cs:0.42, 0.51) {Naive CoT};
\node[red!70!black, font=\sffamily\scriptsize\itshape,
      align=left, anchor=north]
    at (axis cs:0.42, 0.56) {$-75.7$\,pp recall};
\fill[accent!10, opacity=0.5] (axis cs:0.8, 0.8) rectangle (axis cs:1, 1);
\node[accent!70!black, font=\sffamily\tiny\itshape]
    at (axis cs:0.9, 0.95) {ideal};
\end{axis}
\end{tikzpicture}